\documentclass{bmvc2k}
\usepackage[utf8]{inputenc}
\usepackage{amsmath, amsfonts, amssymb}
\usepackage{graphicx}
\usepackage{todonotes}
\usepackage[section]{placeins}
\usepackage[activate={true,nocompatibility},final,tracking=true,kerning=true,spacing=true,factor=1100,stretch=15,shrink=15]{microtype}

\title{Weakly-Supervised 3D Pose Estimation from a Single Image using Multi-View Consistency}

\addauthor{Guillaume Rochette}{g.rochette@surrey.ac.uk}{1}
\addauthor{Chris Russell}{chris.russell@surrey.ac.uk}{1}
\addauthor{Richard Bowden}{r.bowden@surrey.ac.uk}{1}

\addinstitution{
 Centre for Vision, Speech and Signal Processing\\
 University of Surrey\\
 Guildford, UK
}

\runninghead{Rochette, Russell, Bowden}{Weakly-Supervised Pose Estimation with Multi-View Consistency}

\def\eg{\emph{e.g}\bmvaOneDot}

\def\etal{\emph{et al}\bmvaOneDot}

\begin{document}

\maketitle

\begin{abstract}
We present a novel data-driven regularizer for weakly-supervised learning of 3D human pose estimation that eliminates the drift problem that affects existing approaches. We do this by moving the stereo reconstruction problem into the loss of the network itself. This avoids the need to reconstruct 3D data prior to training and unlike previous semi-supervised approaches, avoids the need for a warm-up period of supervised training. The conceptual and implementational simplicity of our approach is fundamental to its appeal. Not only is it straightforward to augment many weakly-supervised approaches with our additional re-projection based loss, but it is obvious how it shapes reconstructions and prevents drift. As such we believe it will be a valuable tool for any researcher working in weakly-supervised 3D reconstruction. Evaluating on Panoptic, the largest multi-camera and markerless dataset available, we obtain an accuracy that is essentially indistinguishable from a strongly-supervised approach making full use of 3D groundtruth in training.
\end{abstract}

\section{Introduction}
\label{sec:Introduction}
Human pose estimation is the task of predicting the body configuration of one or several humans in a given image or sequence of images.
Human poses can be estimated in either 2D or 3D.  Human Pose Estimation in 2D is a trivial task for humans despite dealing with problems such as occlusion and lighting or multiple people in a scene. 3D Human Pose Estimation has to deal with the aforementioned problems for 2D Estimation, but rather than locating the various body parts in the original image, it estimates their 3D position. This adds additional complexity as depth estimation has perspective ambiguities, making the problem hard for humans, even with the benefit of stereo vision. 3D Human Pose Estimation remains a largely unsolved problem, especially with challenging poses or in uncontrolled environments.

Recent approaches, using convolutional neural networks, have achieved impressive results for both 2D and 3D Human Pose Estimation. But deep learning models require ever increasing amounts of data to yield optimal results. For 2D Human Pose Estimation, there exists large high-quality datasets, annotated by crowdsourcing, in uncontrolled environments. This enables the training of accurate and robust models. But for 3D Human Pose Estimation, such large-scale in-the-wild datasets do not yet exist, which is mostly due to the methods used to generate 3D annotation. There are three main sources of 3D data: (1) Synthetic data: which despite the ever increasing realism in terms of graphics and textures, the poses are generated from parametric models handcrafted by humans and therefore do not fully reflect the variability of natural data; (2) Mo-Cap: where the common data acquisition protocol involves the use of markers and specialized sensors in highly controlled environments; and (3) Reconstructed data from multiple 2D detections: which achieves remarkable precision, but involves large numbers of cameras in order to solve joint occlusion phenomena as well as provide high reconstruction quality. This also typically limits the environment to indoor studios.

In this paper, we propose a weakly-supervised approach to train models to regress a 3D Pose from a single 2D Pose using Multi-View Consistency.
To evaluate our approach we conduct a comparative experiment between strongly-supervised and weakly-supervised methods.
In the context of 3D Pose Estimation, we consider a method strongly-supervised if it uses 3D Pose data as groundtruth, while a weakly-supervised approach makes use of weaker forms of data, including 2D Poses and camera calibration.
Our weakly-supervised solution yields comparable results to its strongly-supervised counterpart, while making no use of 3D groundtruth and offers the potential for future training on less constrained data.

\section{Literature Review}
\label{sec:LiteratureReview}
There has been significant recent interest in Human Pose Estimation, in no small part due to its importance in applications such as pedestrian detection, human behaviour understanding and HCI.
We give a brief overview of the fields of both 2D and 3D Human Pose Estimation.

\paragraph{2D Pose Estimation:} \label{subsec:2D} Early approaches involve the extraction of features from the image, followed by a regression or a model fitting step. One such example is Pictorial Structures \cite{Felzenszwalb2005} which consists of finding the optimal locations of body parts in the image by simultaneously minimizing the degree of mismatch of the part in the image, and the degree of deformation of the kinematic model between two parts.

More recently, Convolutional Pose Machines have become a popular approach \cite{Wei2016}. Reusing the Pose Machines concept of Ramakrishna \etal \cite{Ramakrishna2014}, they iteratively infer joint heatmaps, where each new prediction refines the previous inference. This architecture implicitly learns long-range dependencies and multi-part cues, enabling fine-grained joint locations even when dealing with various kinds of occlusion. Stacked Hourglass Networks \cite{Newell2016a}, another popular approach, uses this same concept of iterative refinement of the predictions, but it also includes residual convolutional layers \cite{He2015} and skip connections \cite{Ronneberger2015}.

Further improvements in 2D Human Pose Estimation centered around producing more complex models, either by learning compositional human body models \cite{Sun2017, Tang}, or evaluating at various scales \cite{Ke2018}. Simon \etal \cite{Simon2017} trained models for Hand Pose Estimation with very little annotated data, using a multi-view set-up along with 3D reconstruction and bootstrapping in order to iteratively label data without supervision. Pose Estimation in scenes containing multiple people also represents a challenge, and Cao \etal \cite{Cao2017} presented a framework that first estimates the locations of the body parts using a modified Convolutional Pose Machines architecture, before solving the matching problem using their predicted Part-Affinity-Fields.

\paragraph{3D Pose Estimation:} \label{subsec:3D} While much research on 2D Pose Estimation has focused on dealing with single or sequences of monocular RGB images, the nature of the input for 3D Estimation is more varied including, RGB, RGB-D (captured with devices such as the Kinect \cite{Shotton2011}) or multiple images from calibrated cameras, \eg Human3.6M \cite{Ionescu2014} or Panoptic Studio \cite{Joo2016}.

3D Human Pose Estimation approaches can be categorized as approaches inferring pose directly from an image with their model trained in an end-to-end fashion or approaches that decouple the localization of the 2D landmarks from the 3D lifting step. We can also categorize them by those that make extensive use of 3D annotations, known as strongly-supervised approaches, or make use of weak-supervision to reduce the dependency on 3D annotated data.

Strongly-supervised end-to-end techniques often make use of 2D cues to improve their performance.
Li and Chan \cite{Li2015} proposed a framework that jointly performs a coarse 2D part detection coupled with a 3D Pose regression.
Park \etal \cite{Park2016} improved on the previous solution, incorporating implicit limb dependencies.
Improving further on the part dependencies, Zhou \etal \cite{Zhou2016} introduced Kinematic layers which force the model to produce physically plausible poses.
Using more conventional layers, Sun \etal \cite{Sun2017} designed a compositional loss function, which gives structure-awareness to the network.
Pavlakos \etal \cite{Pavlakos2016} developed a fully convolutional end-to-end architecture, inspired by \cite{Newell2016a}, which discretizes the 3D space and performs a voxel-by-voxel classification.
Tekin \etal \cite{Tekin2017} showed that fusing 2D body part heatmaps produced by a Stacked Hourglass network \cite{Newell2016a} at multiple stages improved performance.

Decoupled approaches make use of advances in 2D localization to estimate 3D Pose.
Chen and Ramanan \cite{Chen2016} presented an example-based method, inspired by \cite{Ramakrishna2012}, which matches 2D Poses produced by an off-the-shelf detector \cite{Wei2016} with a set of 3D Poses, by minimizing the re-projection error.
Bogo \etal \cite{Bogo2016} proposed a solution to fit the parametrizable 3D body shape model SMPL \cite{Loper2015} to 2D landmarks inferred by DeepCut \cite{Pishchulin2015}.
Martinez \etal \cite{Martinez2017} presented a simple yet effective discriminative approach, where a small fully connected network with residual connections is trained to regress 3D Poses from 2D detections produced by a Stacked Hourglass model, and outperformed more complex state-of-the-art approaches. 

All previously discussed supervised methods leverage 3D Poses from groundtruth. Large scale datasets providing accurate 3D groundtruth use either Mo-Cap data (\eg HumanEva \cite{Sigal2006} or Human3.6M \cite{Ionescu2014}) or using a high number of cameras (such as the Panoptic Studio \cite{Joo2016}). These widely used datasets were captured in  controlled environments, which therefore limits the generalization capabilities of the models to "in the wild" images. To overcome this limitation, recent approaches are now aiming at learning 3D Human Pose Estimation using less constraining sources of data, hence named weakly-supervised architectures.

Tome \etal \cite{Tome2017} proposed an architecture, derived from the successful Pose Machines framework \cite{Ramakrishna2014, Wei2016}. It iteratively predicts 2D landmarks, lifts the pose to 3D by finding the best rotation and pose that minimizes the re-projection loss, before fusing 2D and 3D cues to produce a refined 2D pose, and back-propagating solely on the 2D losses.
Rhodin \etal \cite{Rhodin2018a} suggested an end-to-end approach using unlabelled images from multiple calibrated cameras and a small amount of 3D labelled data. They demonstrated that using geometric constraints could extend the generalization capabilities to reliably predict poses in uncontrolled environments.
Inspired by the architecture of Martinez \etal \cite{Martinez2017}, Drover \etal \cite{Drover} trained a generative adversarial network, which from an input 2D pose generates its corresponding 3D pose, with constraints on the depth, before randomly projecting it onto a 2D image plane. The discriminator, given real 2D pose, needs to determine if the projected pose is valid or not.

We aim to learn the mapping from the 2D to 3D Pose distribution space in a weakly-supervised manner, i.e. without any explicit prior on the 3D Pose distribution space or supervised training while avoiding drift.

\section{Weakly-Supervised Learning of 3D Pose Estimation from a Single 2D Pose using Multi-View Consistency}
\label{sec:2D23D}

To train a 2D-to-3D model we use a combination of two weakly-supervised losses, the first enforces a multi-view consistency constraint derived from the images, whereas the second applies a re-projection consistency of the predicted output on its input.
This loss combination enables the use of a calibrated multi-camera set-up that yields images, from which we infer the 2D Poses, from any state-of-the-art 2D Human Pose Estimation framework.

\begin{figure}[h]
\centering
\begin{minipage}{.49\textwidth}
  \centering
  \includegraphics[width=0.8\linewidth]{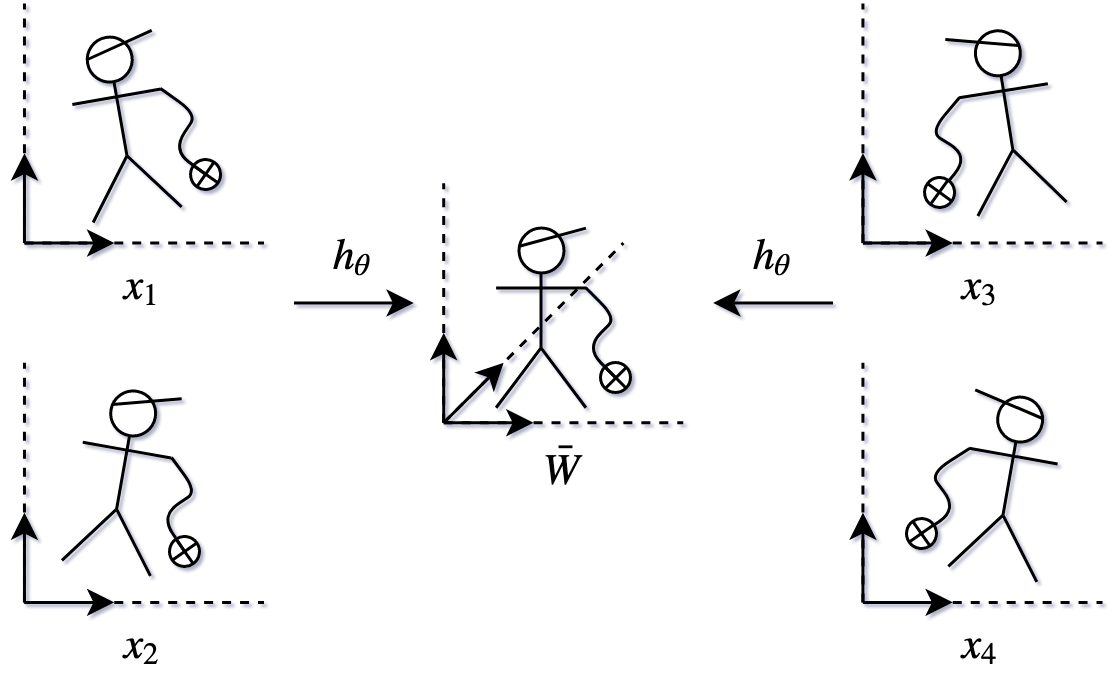}
  \label{fig:multiview_loss}
\end{minipage}%
\unskip\ \vrule \ 
\begin{minipage}{.49\textwidth}
  \centering
  \includegraphics[width=0.8\linewidth]{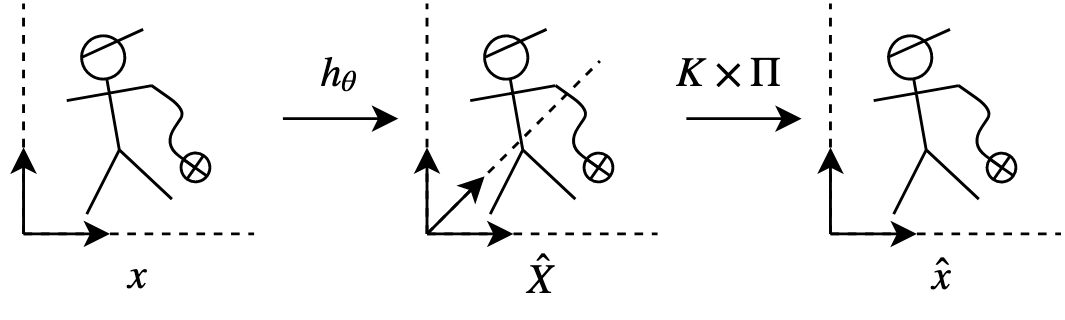}
  \label{fig:re-projection_loss}
\end{minipage}
\caption{Left: Multi-View Consistency enforces of the superimposability of the 3D Poses. Right: Re-Projection Consistency enforce the back-projectability of the 3D Pose into its input 2D Pose.}
\end{figure}

We make the distinction between absolute and relative pose. 
An \emph{absolute} pose, noted $P^a$ is a pose where the origin of the coordinate system is not determined by the location of a root joint. Conversely, a \emph{relative} pose, noted $P^r$, is a pose where an arbitrarily chosen root joint is chosen as the origin of the coordinate system. With the $\alpha$-th joint as the root joint, we have $P^r = P^a - P^a_\alpha$.

Let, $h_\theta$, be a mapping function parameterized by $\theta$, which from a single relative 2D Pose $x^p$ produces a relative 3D Pose $X^p$, such that,
\begin{equation}
    h_\theta : x^r \in \mathbb{R}^{N_J \times 2} \rightarrow X^r \in \mathbb{R}^{N_J \times 3}
\end{equation}
where $N_J$ denotes the number of joints used by our body model.

We define $\mathbf{F}$, as a dataset $\mathbf{F} = {\{f_i\}}_{i=1}^{N}$ of $N$ multi-view frames, where a multi-view frame $f$ is given by,
\begin{equation}
\label{eq:frame}
    f = \{{(I_i, x_i, R_i, t_i, K_i)}_{i=1}^{N_C}\}
\end{equation}
where $N_C$ denotes the number of cameras available for the frame, $I_i \in \mathbb{R}^{H \times W \times C}$ is the image taken from the $i$-th camera, $x_i \in \mathbb{R}^{N_J \times 2}$ is the pose inferred from $i$ by any \emph{state-of-the-art} 2D Human Pose Model $g_\psi$, and $R_i \in \mathbb{R}^{3\times 3}$, $t_i \in \mathbb{R}^{3}$, $K_i \in \mathbb{R}^{3\times 3}$ are respectively the camera calibration parameters of the $i$-th camera.

To train our model, we minimize, with respect to the model parameters $\theta$, the following loss function,
\begin{equation} \label{eq:general_loss}
    \min_\theta L(\mathbf{F}) = \lambda \cdot L_M(\mathbf{F}) + (1 - \lambda) \cdot L_R(\mathbf{F})
\end{equation}
Where, $L_M$, described in \ref{subsec:multiview_loss}, is a loss that enforces a multi-view consistency constraint for the 3D Poses, and $L_R$, described in \ref{subsec:re-projection_loss}, enforces a re-projection constraint on the 3D Pose projected back to its input 2D Pose, and $\lambda \in [0, 1]$, a coefficient balancing the losses.

\subsection{Multi-View Consistency Loss}
\label{subsec:multiview_loss}

For each camera we infer 3D Poses using our model $h_\theta$ from 2D Poses, produced by a \emph{state-of-the-art} 2D Human Pose Model $g_\psi$. These independent 3D Poses should agree when transformed in a unified world coordinate systems using the camera extrinsic parameters, as seen in Fig \ref{fig:multiview_loss}. We can therefore compute, $L_M$, a multi-view consistency loss, that penalizes inconsistencies between the predicted 3D Poses.

We compute unified world-view 3D Poses for every available view, as follows,
\begin{equation}
    \hat{X}^r_i = h_\theta(x^r_i)
\end{equation}
\begin{equation}
    \hat{W}^r_i = R_i^\intercal \, \hat{X}^r_i
\end{equation}
Therefore we can derive the average 3D pose across all views, using the mean as an estimator,
\begin{equation} \label{eq:mean}
    \bar{W}^r = \frac{1}{N_C} \sum_{i=1}^{N_C}{\hat{W}^r_i}
\end{equation}
We can now compute $L_M$, by comparing the average 3D pose $\bar{W}$, as the target, against $\hat{W}$, as the predictions,
\begin{equation} \label{eq:L_M}
    L_M = \sum_{i=1}^{N_C}{\sum_{j=1}^{N_J}{||\bar{W}^r_{j} - \hat{W}^r_{ij}||}}
\end{equation}

Learning using only this loss function does not lead to good quality reconstructions. In particular, it does not penalize the model for producing consistent estimates unrelated to the 2D inputs, it merely ensures that estimates from the different views are consistent. If only this loss is applied, the model will drift and infer constant poses - typically all joints will be clustered on a single central point regardless of the input.

To avoid this, we impose additional data-driven constraints for training, that enforce that the 3D reconstruction is consistent with the 2D input data. In particular we enforce that the re-projection of the 3D pose through the camera matrix is as close as possible to its input 2D pose.

\subsection{Re-Projection Consistency Loss}
\label{subsec:re-projection_loss}

As mentioned in \ref{subsec:multiview_loss}, the Multi-View Consistency Loss does not prevent model drift. In previous work, Rhodin \etal \cite{Rhodin2018a} used a small amount of supervised data, \eg where the 3D pose groundtruth exists, along with a regularization parameter penalizing the model if its predictions were too far from the prediction made by a model trained only on supervised data.
Our contribution overcomes this problem with some simple reasoning.

With 2D poses produced by a \emph{state-of-the-art} 2D Human pose Model $g_\psi$, the 3D Pose inferred by our model $h_\theta$ should also be consistent with input, the 2D Pose, when back-projected, as seen in Figure \ref{fig:re-projection_loss}.

To improve performance and stability, rather than simply re-projecting the inferred 3D Pose, we re-project the averaged 3D relative Pose $\bar{W}^r$ from Eq. \ref{eq:mean}.
We produce the averaged 3D absolute pose using the absolute coordinates of the root joint $W^a_\alpha$, which can either be inferred by an off-the-shelf depth predictor or reconstructed using multiple views,
\begin{equation}
    \bar{W}^a = \bar{W}^r + W^a_\alpha
\end{equation}
and now re-project this averaged 3D Pose into every view, using $\Pi : \mathbb{R}^3 \rightarrow \mathbb{R}^2$, 
\begin{equation}
    \bar{X}^a_i = R_i \, \bar{W}^a + t_i
\end{equation}
\begin{equation}
    \bar{x}^a_i = K_i \, \Pi(\bar{X}^a_i)
\end{equation}
We can now compute $L_R$, by comparing input 2D poses $x$, as the target, against $\bar{x}$, as the prediction,
\begin{equation} \label{eq:L_R}
    L_R = \sum_{i=1}^{N_C}{\sum_{j=1}^{N_J}{||x^a_{ij} - \bar{x}^a_{ij}||}}
\end{equation}

This loss function alone is not sufficient as it cannot constraint the model's depth, but it enforces that the back-projection of the 3D output resembles its 2D input.

We will now describe the initialization issues related to the choice of the projection operator, and its inherent convexity.

\subsubsection{Initialization: Tackling the Non-Convexity of the Perspective Projection}
\label{subsubsec:initialization}
Let, $\Pi_p: \mathbb{R}^3 \rightarrow \mathbb{R}^2$, the perspective projection be given by,
\begin{equation}
    \Pi_p(X) = \frac{1}{X_z}
    \begin{pmatrix}
    X_x \\
    X_y
    \end{pmatrix}
\end{equation}
This hyperbolic function is non-convex, and its Jacobian matrix, is given by,
\begin{equation}
    J_{\Pi_p}(X) =
    \begin{pmatrix}
    \frac{1}{X_z} & 0 & \frac{-X_x}{{(X_z)}^2}\\
    0 & \frac{1}{X_z} & \frac{-X_y}{{(X_z)}^2}
    \end{pmatrix}
\end{equation}
Depending of the magnitude of $X_z$, it can lead to either exploding or vanishing gradients.
Therefore, without proper initialization of the model's parameters, it is unlikely that the model will converge. This is why, for the early stage of the training, we replace the perspective projection by the orthographic projection, $\Pi_o: \mathbb{R}^3 \rightarrow \mathbb{R}^2$, which is a linear function and therefore convex, given by,
 \begin{equation}
    \Pi_o(X) =
    \begin{pmatrix}
    X_x \\
    X_y
    \end{pmatrix}
\end{equation}

\subsection{Panoptic Studio Dataset}
\label{subsec:panoptic_dataset}
We used the Panoptic Studio Dataset of Joo \etal \cite{Joo2016}, which provides high fidelity 3D Poses produced in a markerless fashion. There are over 70 sequences, captured from multiple cameras: 480 VGA cameras, 31 HD cameras and 10 Kinects for depth point clouds. Some sequences include multiple persons engaged in social activities, such as a band playing music, or a crowd playing role playing games. Other sequences focus on single individuals dancing or simply posing.
A complex methodology explained in detail in \cite{Joo2016}, which includes robust 3D reconstruction from many 2D detections and the use of temporal cues, enables the production of very high fidelity groundtruth, which for evaluation we will consider as the most accurate existing 3D estimates for the dataset.

To avoid the difficulties of matching people across different viewpoints and to minimize issues with occlusion, we restricted ourselves to the recent Panoptic single person sequences. The Body Poses are represented with 18 joints in the COCO format, and the Hand Poses are described with 21 keypoints per hand.

As for the separation of the data into training, validation and test set, we split such that entire sequences are in one of the three sets as a whole, aiming roughly at $80\%$ for training, $10\%$ for validation and $10\%$ for testing. The properties of the dataset are described in Table \ref{tab:panoptic_dataset}.

\begin{table}[h]
\centering
\small\addtolength{\tabcolsep}{-5pt}
\begin{tabular}{|l|ccc|}
\hline
Set & Frames & Views & Individuals \\ \hline
Training & 220553 & 18-31 & 40 \\
Validation & 27575 & 31 & 6 \\
Test & 25366 & 31 & 6 \\ \hline
Total & 273494 & - & 52 \\ \hline
\end{tabular}
\caption{Description of the Panoptic Dataset}
\label{tab:panoptic_dataset}
\end{table}

\subsection{Experimental Setup and Results}
\label{subsec:experiments}
We compare the performance between strongly-supervised and weakly-supervised training schemes, as the measure of a good weakly-supervised approach is that it should yield performance as close as possible to a strongly-supervised approach. We therefore present two comparative studies.

The experiments make use of 2D Detections, obtained with the state-of-the-art 2D Detector, OpenPose \cite{Cao2017}, which uses the Convolutional Pose Machines introduced by Wei \etal \cite{Wei2016} as well as Simon \etal \cite{Simon2017} for Hand Poses.
For the first experiment, we focused on training models to predict body joint positions.
Whereas for the second experiment, we trained models to predict body and hand joint positions.

Using 3D Panoptic Poses as groundtruth for training the strongly-supervised models would not be fair, as it was reconstructed using temporal relationships between frames, and our weakly-supervised experiment does not make use of any temporal cues whatsoever. We therefore reconstruct 3D Poses from the detected 2D Poses by implementing our own 3D reconstruction method without temporal information. We follow the two-step approach detailed by Faugeras \cite{Faugeras1993}, which first computes a closed-form solution using the orthographic projection, followed by an iterative method using the perspective projection. The weakly-supervised and strongly-supervised experiments have exactly the same architectures and hyper-parameters.

\subsubsection{Implementation}
\begin{figure}[h]
    \centering
    \includegraphics[width=.9\linewidth]{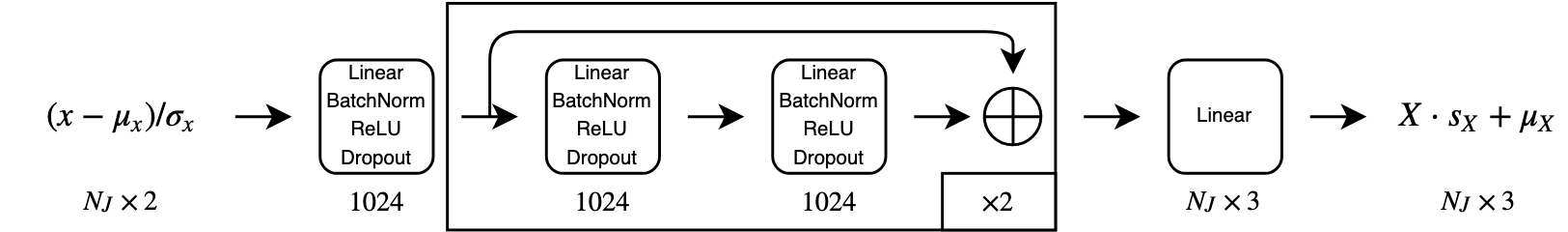}
    \caption{Diagram of the mapping model inferring a 3D Pose from a single 2D Pose.\label{fig:model}}
\end{figure}

We implemented a slightly modified version of the model introduced by Martinez \etal \cite{Martinez2017},  consisting of 6 Linear Layers, each followed by a Batch Normalization \cite{Ioffe2015}, ReLU \cite{Nair2010} and Dropout \cite{Srivastava2014} layers, with residual connections, as seen in Fig \ref{fig:model}. We decrease the learning rate every epoch using Exponential Decay. Rather than clipping the model parameters, we use a simple weight decay as a regularization term.
Our model was trained to handle incomplete 2D pose input, as it was trained using 2D poses with missing joints.
The network takes as input the 2D Pose relative to a given root joint, given by, $x^r_N = \frac{x^r - \mu_x}{\sigma_x}$, with the mean $\mu_x$ and the standard deviation $\sigma_x$. It also produces as 3D output a root joint and its normalized relative joints, which we unnormalize as follows, $X^r = X^r_N \cdot \sigma_X + \mu_X$.  ($\mu_x, \sigma_x$) and ($\mu_X, \sigma_X$) were computed over the training set. 
Training was done in batches of frames, for which a multi-view frame is given by, $f = \{{(x_i, R_i, t_i, K_i)}_{i=1}^{N_C}\}$.

\subparagraph{Training Details}
We trained both the strongly-supervised and the weakly-supervised models for $100$ epochs with the following hyper-parameters: $1024$ units for every hidden layer using Xavier initialization \cite{Glorot2010}, $8$ frames of $16$ views per batch, Adam \cite{Kingma2014} with the learning rate $\alpha=5 \cdot {10}^{-4}$, and the exponential decay $\gamma=0.96$.

Concerning the weakly-supervised model, the balancing coefficient $\lambda=0.8$ in Eq. \ref{eq:general_loss}.

The models were trained to minimize the Huber Loss over all terms indicated by the norm symbol $|| \cdot ||$, \eg Eq. \ref{eq:L_M} \& \ref{eq:L_R}. The Huber loss is defined as:
\begin{equation}
    L(x, y) = 
    \begin{cases}
        \frac{1}{2}{|x - y|}^2 & \textrm{if } |x - y| \leq \delta, \\
        \delta (|x - y| - \frac{1}{2} \delta) & \textrm{otherwise.}
    \end{cases}
\end{equation}
The Huber Loss with $\delta = 1$, known as Smooth $L_1$ Loss, was chosen over $L_1$ and $L_2$ losses, as it was beneficial for both strongly-supervised and weakly-supervised methods.

\subsubsection{Results and Analysis}
We evaluated our models by measuring the average distance on a per-joint basis between the predicted pose $\hat{X}$ and its groundtruth $X$, without any post-processing, such that,
\begin{equation}
    \label{eq:distance}
    \Delta(\hat{X}^r, X^r) = \frac{1}{N_J} \sum_{j=1}^{N_J}{{||\hat{X}^r_j - X^r_j||}_2}
\end{equation}
We use early-stopping by looking at the validation error, and report the results on the test set. Accuracy is compared against the Panoptic groundtruth.

\begin{table}[h]
    \centering
    \begin{minipage}{.49\textwidth}
        \scriptsize\addtolength{\tabcolsep}{-5pt}
        \centering
        \begin{tabular}{l|cc|}
        \cline{2-3}
        & \multicolumn{1}{l}{Strongly-Supervised} & \multicolumn{1}{l|}{Weakly-Supervised} \\ \hline
        \multicolumn{1}{|l|}{Body} & 71.013 & 71.019 \\
        \multicolumn{1}{|l|}{Body and Hands} & 84.115 & 87.015 \\ \hline
        \end{tabular}
        \caption{Results of the average error in \emph{mm} on the test set.}
        \label{tab:results}
    \end{minipage}%
    \unskip\ \vrule \ 
    \begin{minipage}{.49\textwidth}
        \tiny\addtolength{\tabcolsep}{-5pt}
        \centering
        \begin{tabular}{lcclcclll}
        \cline{2-3} \cline{5-6} \cline{8-9}
        \multicolumn{1}{l|}{} & S.-S. & \multicolumn{1}{c|}{W.-S.} & \multicolumn{1}{l|}{} & S.-S. & \multicolumn{1}{c|}{W.-S.} & \multicolumn{1}{l|}{} & S.-S. & \multicolumn{1}{l|}{W.-S.} \\ \hline
        \multicolumn{1}{|l|}{Nose} & \textbf{28.57} & \multicolumn{1}{c|}{29.02} & \multicolumn{1}{l|}{L. Elbow} & 57.60 & \multicolumn{1}{c|}{\textbf{55.10}} & \multicolumn{1}{l|}{L. Knee} & \textbf{60.68} & \multicolumn{1}{l|}{61.49} \\
        \multicolumn{1}{|l|}{Neck} & - & \multicolumn{1}{c|}{-} & \multicolumn{1}{l|}{L. Wrist} & 85.19 & \multicolumn{1}{c|}{\textbf{82.82}} & \multicolumn{1}{l|}{L. Ankle} & \textbf{97.67} & \multicolumn{1}{l|}{100.15} \\
        \multicolumn{1}{|l|}{R. Shoulder} & \textbf{18.80} & \multicolumn{1}{c|}{19.20} & \multicolumn{1}{l|}{R. Hip} & \textbf{40.31} & \multicolumn{1}{c|}{40.82} & \multicolumn{1}{l|}{R. Eye} & \textbf{57.45} & \multicolumn{1}{l|}{59.02} \\
        \multicolumn{1}{|l|}{R. Elbow} & 59.52 & \multicolumn{1}{c|}{\textbf{57.60}} & \multicolumn{1}{l|}{R. Knee} & \textbf{60.95} & \multicolumn{1}{c|}{61.95} & \multicolumn{1}{l|}{L. Eye} & 77.59 & \multicolumn{1}{l|}{\textbf{77.46}} \\
        \multicolumn{1}{|l|}{R. Wrist} & 85.40 & \multicolumn{1}{c|}{\textbf{84.01}} & \multicolumn{1}{l|}{R. Ankle} & 105.88 & \multicolumn{1}{c|}{\textbf{105.87}} & \multicolumn{1}{l|}{R. Ear} & \textbf{153.20} & \multicolumn{1}{l|}{154.08} \\
        \multicolumn{1}{|l|}{L. Shoulder} & \textbf{18.96} & \multicolumn{1}{c|}{19.30} & \multicolumn{1}{l|}{L. Hip} & 38.90 & \multicolumn{1}{c|}{\textbf{38.64}} & \multicolumn{1}{l|}{L. Ear} & \textbf{160.55} & \multicolumn{1}{l|}{160.80} \\ \hline
        \end{tabular}
        \caption{Detailed results in \emph{mm} on a per-joint basis for body pose models.}
        \label{tab:body_results}
    \end{minipage}
    \label{tab:my_label}
\end{table}

\begin{figure}[h]
\centering
\begin{minipage}{.9\textwidth}
  \centering
  \includegraphics[width=\linewidth]{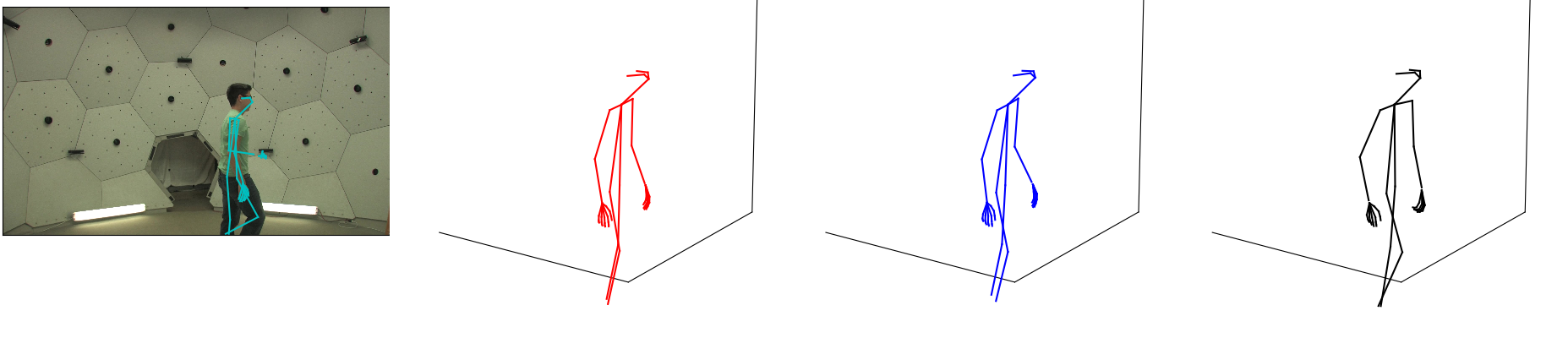}
\end{minipage}

\begin{minipage}{.5\textwidth}
  \centering
  \includegraphics[width=\linewidth]{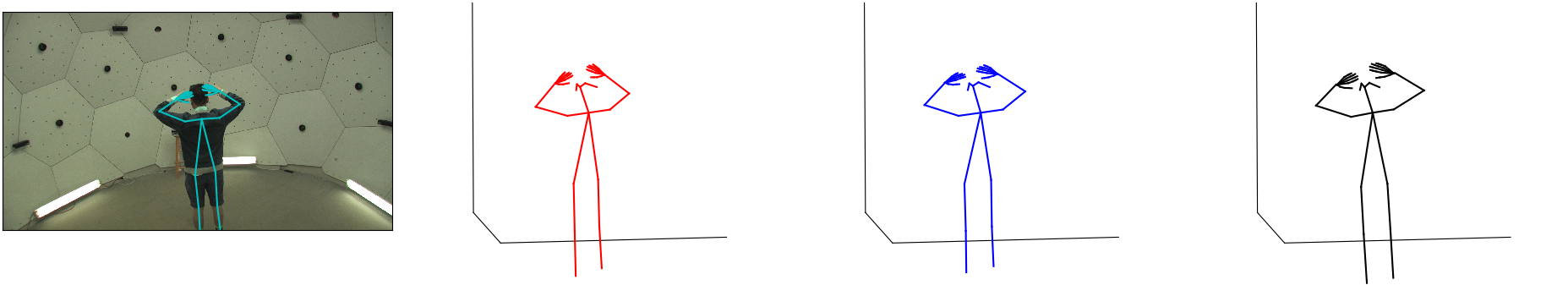}
\end{minipage}%
\begin{minipage}{.5\textwidth}
  \centering
  \includegraphics[width=\linewidth]{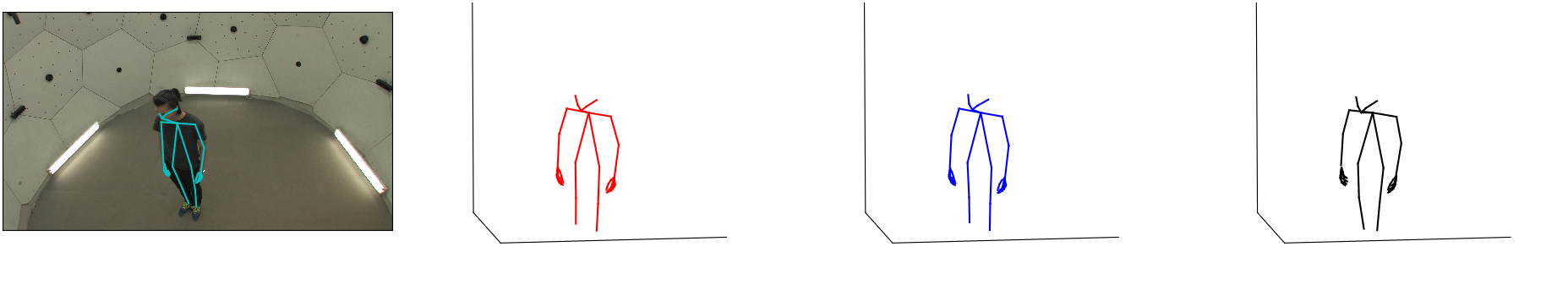}
\end{minipage}

\begin{minipage}{.5\textwidth}
  \centering
  \includegraphics[width=\linewidth]{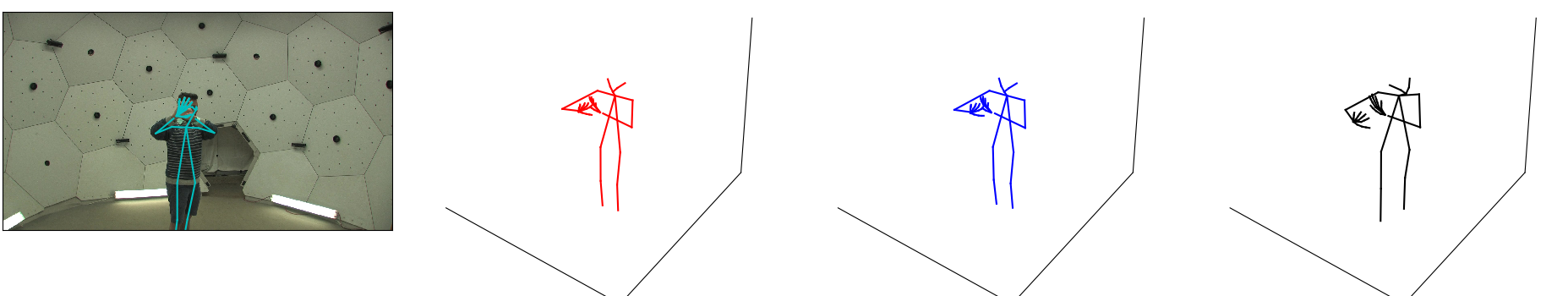}
\end{minipage}%
\begin{minipage}{.5\textwidth}
  \centering
  \includegraphics[width=\linewidth]{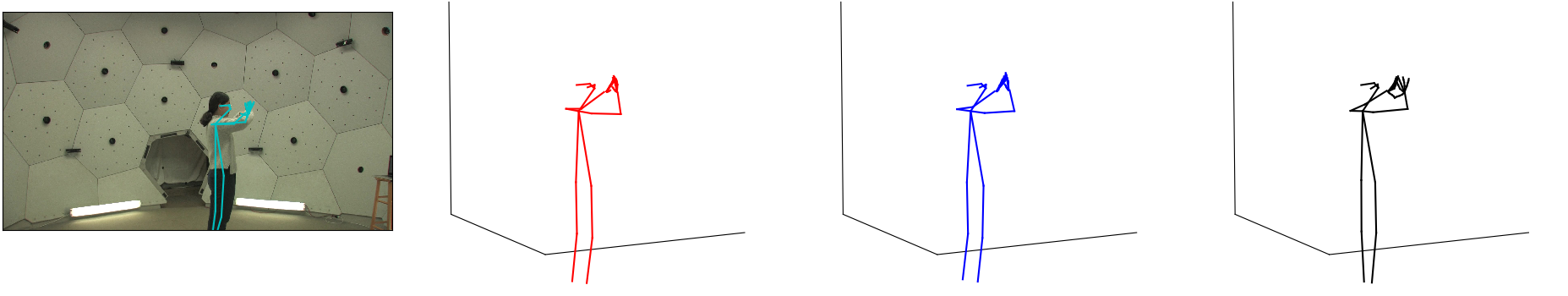}
\end{minipage}

\begin{minipage}{.5\textwidth}
  \centering
  \includegraphics[width=\linewidth]{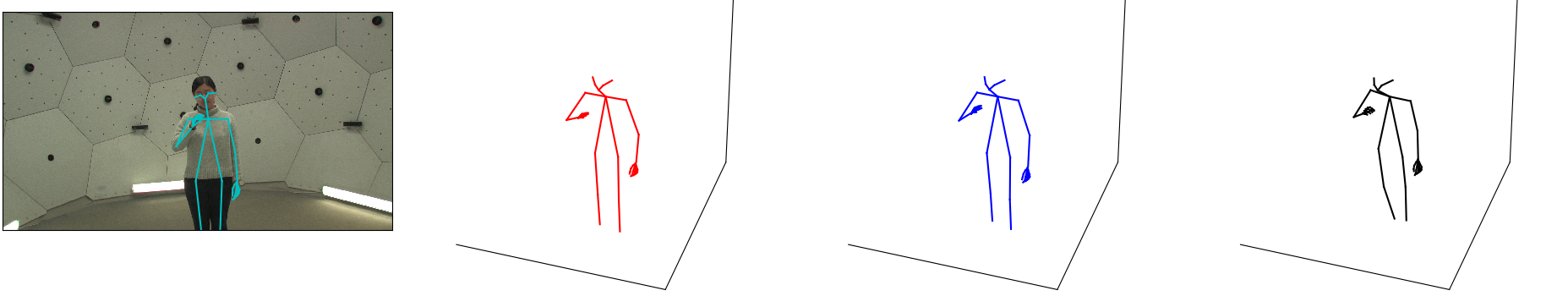}
\end{minipage}%
\begin{minipage}{.5\textwidth}
  \centering
  \includegraphics[width=\linewidth]{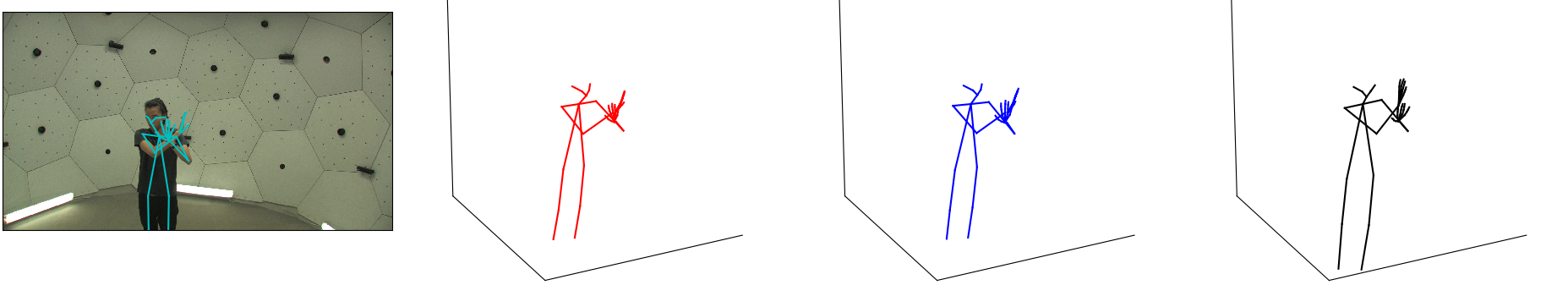}
\end{minipage}
\caption{Examples on the test set of Panoptic Dataset. From Left to Right: (1) Image with OpenPose 2D Detection. (2) Strongly-Supervised Prediction. (3) Weakly-Supervised Prediction. (4) Panoptic 3D Groundtruth.}
\label{fig:examples}
\end{figure}

\begin{figure}[h]
\centering
\begin{minipage}{.9\textwidth}
  \centering
  \includegraphics[width=\linewidth]{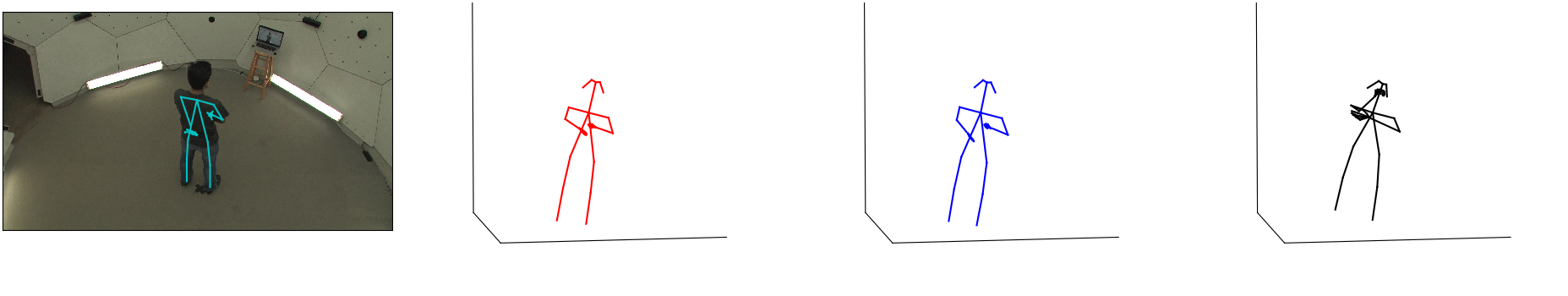}
\end{minipage}
\begin{minipage}{.9\textwidth}
  \centering
  \includegraphics[width=\linewidth]{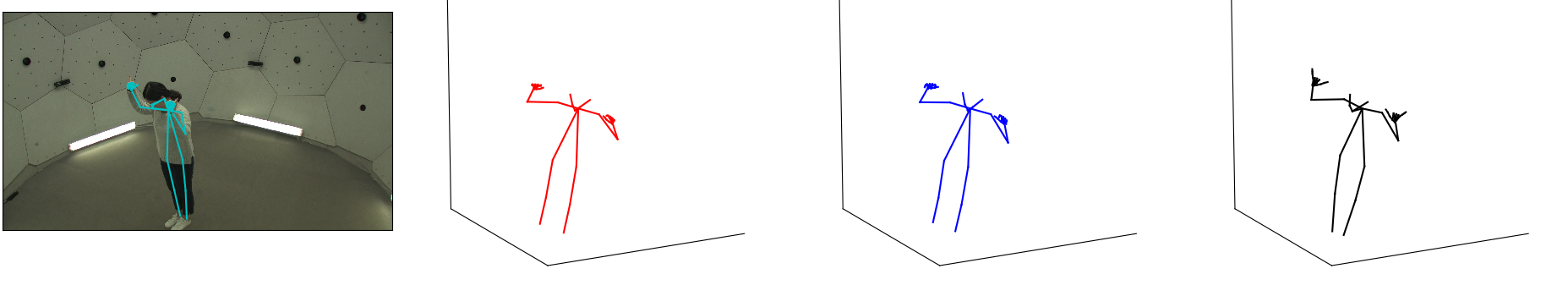}
\end{minipage}
\caption{Failure Cases. Top: OpenPose mis-detection of the left arm resulted in a coherent yet inaccurate estimation. Bottom: OpenPose partially fails at retrieving the correct 2D Pose, resulting in both models to fail at estimating the 3D hand poses.}
\label{fig:failures}
\end{figure}

Table \ref{tab:results} shows that the weakly-supervised model is equivalent in performance to a strongly-supervised model trained with 3D groundtruth with less than $1/100$ \emph{mm} difference. The average error for the body and hands model had only a $3$ \emph{mm} difference but since the best error were received in respectively the $94$ and $97$ epoch, it is likely that further training might provide further improvement.

Table \ref{tab:body_results} provides detailed results on a per joint basis. Interestingly, the weakly-supervised approach is more accurate at estimating the elbows and wrists which are notoriously more challenging to estimate due to high variability. For both the supervised and weakly-supervised approaches, ankles and ears provide the worst results. This is predominantly due to the fact that OpenPose is particularly poor at detecting the ears in 2D and the ankles are often not visible in the image.

Figure \ref{fig:examples} shows some qualitative examples. As can be seen, both the strongly and weakly-supervised approach closely match the groundtruth.  Figure \ref{fig:failures} shows failure cases where the top example shows a failure due to occlusion and the bottom example is incorrectly estimating the hand due to a rare hand pose which is partially mis-detected. In both cases these issues are manifest in both the strongly and weakly-supervised cases showing that it is not a limitation of the weakly-supervised architecture. 

\section{Conclusion}
Our weakly-supervised formulation to 3D human pose reconstruction makes them directly comparable to a strongly-supervised approach, both in terms of the accuracy of the results and in the elimination of drift resulting in greater stability and convergence. We have presented a semi-supervised approach to human-pose estimation that is essentially indistinguishable in its results from a strongly-supervised approach. 

The conceptual and implementational simplicity of our approach is fundamental to its appeal. Not only is it straightforward to augment many weakly-supervised approaches with our additional re-projection based loss, but it is obvious how it shapes reconstructions and prevents drift. As such we believe it will be a valuable tool for any researcher working in the weakly-supervised 3D reconstruction. With an absence of access to 3D data being one of the fundamental limitations as we move forward in 3D reconstruction, these weakly-supervised approaches will only grow in importance.

\section{Acknowledgements}
This project has received funding from the European Union’s Horizon 2020 research and innovation programme under grant agreement No 762021 (Content4All). This work reflects only the author’s view and the Commission is not responsible for any use that may be made of the information it contains. We would also like to thank NVIDIA Corporation for their GPU grant.
\bibliography{egbib}
\end{document}